%% file: naaclhlt2019.tex
\documentclass[11pt,a4paper]{article}
\usepackage[hyperref]{naaclhlt2019}
\usepackage{times}
\usepackage{latexsym}

\usepackage{url}

\usepackage{float}
\usepackage{subcaption}
\usepackage{graphicx}  
\usepackage{ragged2e}
\usepackage{xcolor}
\usepackage{courier}

\definecolor{cadmiumgreen}{rgb}{0.0, 0.42, 0.24}
\aclfinalcopy 
\def\aclpaperid{993} 

\usepackage{anyfontsize}

\title{Multi-Level Memory for Task Oriented Dialogs}

\author{Revanth Reddy$^1$\thanks{\,  Work done during internship at IBM Research AI}, \quad Danish Contractor$^2$, \quad Dinesh Raghu$^2$, \quad Sachindra Joshi$^2$\\
$^1$Indian Institute of Technology, Madras \qquad $^2$IBM Research AI, New Delhi\\
\texttt{g.revanthreddy111@gmail.com}\\\texttt{\{dcontrac,diraghu1,jsachind\}@in.ibm.com}}

\date{}

\begin{document}
\maketitle
\begin{abstract}
  Recent end-to-end task oriented dialog systems use memory architectures to incorporate external knowledge in their dialogs. Current work makes simplifying assumptions about the structure of the knowledge base (such as the use of triples to represent knowledge) and combines dialog utterances (context), as well as, knowledge base (KB) results, as part of the same memory. This causes an explosion in the memory size, and makes reasoning over memory, harder. In addition, such a memory design forces hierarchical properties of the data to be fit into a triple structure of memory. This requires the memory reader to learn how to infer relationships across otherwise connected attributes.

In this paper we relax the strong assumptions made by existing architectures and use separate memories for modeling dialog context and KB results. Instead of using triples to store KB results, we introduce a novel multi-level memory architecture consisting of cells for each query and their corresponding results. The multi-level memory first addresses queries, followed by results and finally each key-value pair within a result. We conduct detailed experiments on three publicly available task oriented dialog data sets and we find that our method conclusively outperforms current state-of-the-art models. We report a 15-25\% increase in both entity F1 and BLEU scores. 
\end{abstract}

\input{intro_4.tex}
\input{related.tex}
\input{method.tex}

\input{experiments.tex}

\input{conclusion.tex}
\bibliography{naaclhlt2019}
\bibliographystyle{acl_natbib}

\end{document}

%% file: intro_4.tex
\section{Introduction}
Task oriented dialog systems are designed to complete a user specified goal, or service an information request using natural language exchanges. Unlike open domain end-to-end neural dialog models, task oriented systems rely on external knowledge sources, outside of the current conversation context, to return a response ~\cite{DSTC2,CamRest,babI,KVRet,elasri2017frames}. For instance, in the example shown in Table \ref{tab:example}, a dialog agent giving tour package recommendations needs to be able to first query an external knowledge source to determine packages that meet a user's requirement, and then respond accordingly.  


\begin{table}[!ht]
 \centering
 \tiny
 
 \begin{tabular}{lllllll }
\textbf{Origin} & \textbf{Dest. } & \textbf{Hotel}   &  \textbf{Price} &\textbf{Cat.}&\textbf{Rating} & \textbf{...}                                                                                                                                                                                                              \\
\hline \hline 
Dallas & Mannheim & Regal Resort  &\$2800&5.0&8.98& ...\\
Toronto & Calgary & Amusement & \$1864.4 &4.0&6.91& ...\\
Dallas & Mannheim & Vertex Inn  &\$3592.8&3.0&7.15& ...\\
Dallas & Santos & Globetrotter& \$2000 &3.0 &8.37& ..\\
Dallas & Mannheim & Starlight & \$4018.1 & 4.0 & 8.56\\
... & ... & ... & ... &... & ...& ...\\
\hline \hline
\end{tabular}
 
 \begin{tabular}{lll}
\\ \textbf{Role} & \textbf{Turn } & \textbf{Utterance}   \\                                                                                                                                                                                                         
 \hline \hline 
Agent         & 1              & how can i help                                                                                                                                                                                                                        \\ \cline{3-3}
 User          & 2              & \begin{tabular}[c]{@{}l@{}}i've got a few days off~ from~ aug 26-aug 31. I'm not flexible\\ on this, but i want to somehow treat myself with an 8 day trip.\\ I'm leaving Dallas and I want to check out Munich \end{tabular}  \\
 \cline{3-3}
Agent         & 2              & sorry we dont have any package for Munich   \\   \cline{3-3}    
User          & 3              & how about Mannheiem?    
\\ \cline{3-3}
 Agent         & 3              & would a 5 day trip suffice                                                                                                                                                                                                                 \\ \cline{3-3}
 User          & 4              & sure dude                                                                                                                                                                                                                             \\ \cline{3-3}
 Agent         & 4              & \begin{tabular}[c]{@{}l@{}}A 5 star hotel~ called the Regal resort, it has free wifi and a spa\end{tabular}                                                                                                                         \\ \cline{3-3}
 User          & 5              & dates?                                                                                                                                                                                                                                \\ \cline{3-3}
 Agent         & 5              & starts on august 27th until the 30th                                                                                                                                                                                                  \\ \cline{3-3}
 User          & 6              & \begin{tabular}[c]{@{}l@{}}ok that could work. I would like to see my options in Santos.\end{tabular}                                                                                                                         \\ \cline{3-3}
 Agent         & 6              & \begin{tabular}[c]{@{}l@{}}there is Hotel Globetrotter in Santos. it has 3 stars and comes \\ with breakfast and wifi, it leaves on august 25th and returns on \\ 30th, all for \$2000\end{tabular} \\ \cline{3-3}
 User          & 7              & \begin{tabular}[c]{@{}l@{}}ahh i can't leave until august 26th though. I guess i'll just have \\ to go to Mannheim then. \end{tabular}                                                                                                                                                                                          \\ \cline{3-3}
 Agent         & 7              & \begin{tabular}[c]{@{}l@{}}the Regal resort package costs \$2800 with economy class.              \end{tabular}                                                                                                                                                             \\ \cline{3-3}
 User          & 8              & yeah i will book it                                                                                                                                                                                                                   \\
 \hline
 \end{tabular}
 \caption{A goal oriented dialog based on the Frames dataset \cite{elasri2017frames} along with an external knowledge source with each row containing a tour package.} 
 \label{tab:example}
 \end{table}
In order to enable end-to-end goal oriented dialog tasks, current state of the art methods use neural memory architectures to incorporate external knowledge~\cite{CamRest,KVRet,Mem2Seq}. As can be seen in Table \ref{tab:example}, agent responses may also include entity values present only in the dialog context (eg: ``Munich'' in the Agent response in Turn 2). In order to support such utterances, models also include tokens from the input dialog context in the same memory \cite{Mem2Seq}. 

Existing memory based architectures for task oriented dialog suffer from multiple limitations. 
First, the creation of a shared memory for copying values from dialog context, as well as the knowledge base (KB) results, forces the use of a common memory reader for two different types of data. This makes the task of reasoning over memory, harder -- not only does the memory reader need to determine the right entries from a large memory (since each word from context also occupies a memory cell), it also needs to learn to distinguish between the two forms of data (context words and KB results) stored in the same memory.
\begin{table}[ht]
\centering
\tiny
\begin{tabular}{lll||lll }
\textbf{Subject} & \textbf{Relation } & \textbf{Object}   &  \textbf{Subject} &\textbf{Relation}&\textbf{Object}                                                                                                                                                                                                               \\
\hline \hline 
\textcolor{black}{Vertex Inn} & \textcolor{black}{Price} & \textcolor{black}{\$3592.8} &\textcolor{black}{Vertex Inn}& \textcolor{black}{Category} &\textcolor{black}{3.0}\\

\textcolor{blue}{Regal Resort} &\textcolor{blue}{Price}&\textcolor{blue}{\$2800}&\textcolor{black}{Regal Resort} & \textcolor{black}{Rating} & \textcolor{black}{8.98} \\

\textcolor{blue}{Regal Resort} & \textcolor{blue}{Category} & \textcolor{blue}{5.0} &\textcolor{black}{Starlight} &\textcolor{black}{Price} &\textcolor{black}{\$4018.1}\\
\textcolor{black}{Starlight}&\textcolor{black}{Rating}&\textcolor{black}{8.56}&\textcolor{black}{Starlight} & \textcolor{black}{Category} & \textcolor{black}{4.0} \\
... & ... & ... &... & ...& ...\\
\hline \hline
\end{tabular}
\caption{Results from Dallas to Mannheim stored in the form of triples.} 
\label{tab:ex-triples}
\end{table}

Second, all current neural memory architectures store results, returned by a knowledge source, in the form of triples (eg. $subject-relation-object$). This modeling choice makes it hard for the memory reader to infer relationships across otherwise connected attributes. For instance, consider the example triple store in Table \ref{tab:ex-triples} showing results for a query executed for packages between ``Dallas'' and ``Mannheim''. If the user asks the dialog agent to check the price of stay at a 5 star hotel, the memory reader needs to infer that the correct answer is \$2800 by learning that the price, category and hotel need to be linked inorder to return an answer (shown in \textcolor{blue}{blue}).

Lastly, \textcolor{black}{current models treat conversations as a sequential process, involving the use of KB results from only the most recent information request/query. In contrast, in real world dialogs such as the one shown in Table \ref{tab:example},
the agent may have to refer to results (\textit{to Mannheim}) from a previously executed query (see Turn 7). Thus, at each turn, the system has to memorize all the information exchanged during the dialog, and infer the package being referred to, by the user. 
In order to support such dialogs, the memory needs to store results of all queries executed during the course of the dialog. The problem of inferring over such results (which may be from multiple queries) is exacerbated when memory is represented in the form of triples.} 

In this paper, we present our novel multi-level memory architecture that overcomes the limitations of existing methods: (i) We separate the memory used to store tokens from the input context and the results from the knowledge base. Thus, we learn different memory readers for context words as well for knowledge base entities (ii) Instead of using a $subj-rel-obj$ store, we develop a novel multi-level memory architecture which encodes the natural hierarchy exhibited in knowledge base results by storing queries and their corresponding results and values at each level. 
We first attend on the queries, followed by the results in each query to identify the result being referred to, by the user. We then attend on the individual entries in the result to determine which value to copy in the response.
Figure \ref{fig:memory-rep} shows our multi-level memory storing the results from queries executed as part of the dialog in Table \ref{tab:example}.

\noindent Our paper makes the following contributions:

\noindent {\bf 1. } We propose the use of separate memories for copying values from context and KB results. Thus, the model learns separate memory readers for each type of data.

\noindent {\bf 2. } Our novel multi-level memory for KB results, models the queries, results and their values in their natural hierarchy. As our experiments show, the separation of memory as well as our multi-level memory architecture, both, contribute to significant performance improvements.

\noindent {\bf 3. } We present detailed experiments demonstrating the benefit of our memory architecture along with model ablation studies. Our experiments on three publicly available datasets ({\bf CamRest676} \cite{CamRest}, {\bf InCar assistant} \cite{KVRet}, {\bf Maluuba Frames} \cite{elasri2017frames}) show a substantial improvement of 15-25 \%  in both entity F1 scores, and BLEU scores as compared to existing state of the art architectures. To the best of our knowledge, we are the first to attempt end-to-end modeling of task oriented dialogs with non-sequential references as well as multiple queries, as seen in the Maluuba Frames dataset. A human evaluation on model outputs also shows our model is preferred by users over existing systems such as KVRet \cite{KVRet} and Mem2Seq \cite{Mem2Seq}.  

%% file: related.tex
\section {Related work}

Recent methods, such as \cite{seq2seqconv,serban2016building,serban2017hierarchical}, proposed for end-to-end learning of dialogs were aimed at modeling open-domain dialogs. While they can be used for learning task oriented dialogs, they are not well suited to interface with a structured KB. To better adapt them to handle task oriented dialogs: 1) \cite{BordesW16} proposed a memory network based architecture to better encode KB tuples and perform inferencing over them and 2) \cite{Mem2Seq} incorporated copy mechanism to enable copying of words from the past utterances and words from KB while generating responses. 
All successful end-to-end task oriented dialog networks \cite{KVRet,BordesW16,Mem2Seq} make assumptions while designing the architecture: 1) KB results are assumed to be a triple store, 2) KB triples and past utterances are forced to be represented in a shared memory to enable copying over them. Both these assumptions make the task of inferencing much harder. Any two fields linked directly in the KB tuple are now linked indirectly by the subject of the triples. Further, placing the KB results and the past utterances in same memory forces the architecture to encode them using a single strategy. 
In contrast, our work uses two different memories for past utterances and KB results. The decoder is equipped with the ability to copy from both memories, while generating the response. The KB results are represented using a multi-level memory which better reflects the natural hierarchy encoded by sets of queries and their corresponding result sets. 

Memory architectures have also been found to be helpful in other tasks such as question answering. Work such as \cite{HMNCOLING} defines a hierarchical memory architecture consisting of sentence level memory followed by word memory for a QA task while \cite{HMN2016} defines a memory structure that speeds up loading and inferencing over large knowledge bases. Recent work by \cite{WWW2018} uses a variational memory block along with a hierarchical encoder to improve diversity of open domain dialog responses. 

%% file: method.tex
\section{Multi-Level Memory Network}

\begin{figure*}[ht]
     \begin{subfigure}[b]{0.50\textwidth}
      
       \includegraphics[scale=0.45]{figs/Multi_level_memory_colour.png}
       \caption{{\fontsize{10}{12}\selectfont Architecture of our model with multi-level memory attention}}
       \label{fig:arch}
     \end{subfigure}
     \hfill
     \begin{minipage}[b]{0.46\textwidth}
       \begin{subfigure}[b]{0.98\textwidth}
       \hspace{17mm}
       \includegraphics[scale=0.35]{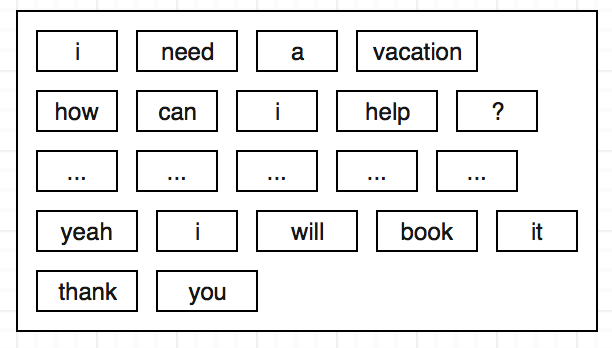}
        
          \caption{{\fontsize{9.38}{12}\selectfont Context memory created using \\the hidden states
          $h^e_{ij}$}}\label{fig:context-mem}
       \end{subfigure}\\
       \begin{subfigure}[b]{\textwidth}
       \includegraphics[scale=0.29]{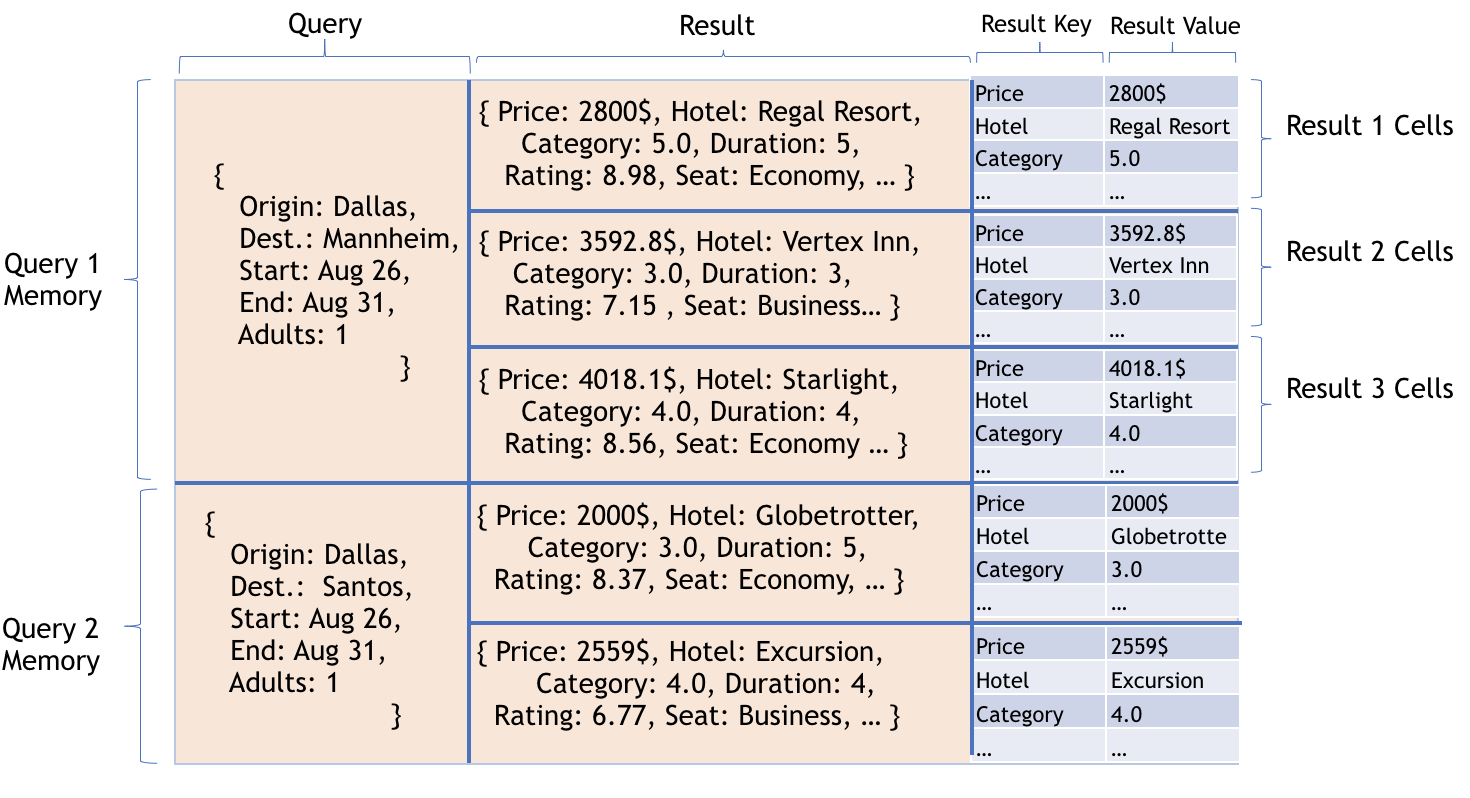}
       
         \caption{{\fontsize{10}{12}\selectfont Expanded view of the multi-level KB memory corresponding to example in Table \ref{tab:example}} }\label{fig:memory-rep}
       \end{subfigure}
     \end{minipage}
     \caption{ Model architecture (a) along with schematic representation of context memory (b) and multi-level KB memory (c)}
\end{figure*}

In this section, we describe our end-to-end model for task oriented dialogues. Our model\footnote{Code is available at \href{https://github.com/DineshRaghu/multi-level-memory-network}{Multi-Level Memory}}  (Figure \ref{fig:arch}) consists of:  (i) a hierarchical encoder which encodes the current input context consisting of the user and agent utterances (ii) a multi-level memory that maintains the queries and knowledge base results seen so far in the course of the dialogue, and (iii) copy augmented sequence decoder that uses separate context and multi-level memory. The queries and their corresponding results are maintained in a multi-level memory. The decoder uses a gating mechanism for memory selection while generating a response. 


\subsection{Encoder} \label{sec:encoder}
Our model uses a standard hierarchical encoder as proposed by \cite{HRED}. The encoder takes a sequence of utterances as input. For the $t^{th}$ turn, the dialogue context can be represented as $(c_1,c_2,...c_{2t-1})$, which consists of $t$ user utterances and $t-1$ system utterances. Each utterance $c_i$ is further a sequence of words $(w_{i1},w_{i2},...w_{im})$. We first embed each word $w_{ij}$ using a word embedding function $\phi^{emb}$ that maps each word to a fixed-dimensional vector. 
We then generate utterance representations, $\varphi(c_i)$ using a single layer bi-directional GRU. \textcolor{black}{$h^e_{ij}$ denotes the hidden state of word $w_{ij}$ in the bi-directional GRU.} 
\textcolor{black}{The input representation $c$ is generated by passing each utterance representation $\varphi(c_i)$ through another single layer GRU.}  

\subsection{Multi-level Memory} \label{sec:memory}


\noindent {\bf Motivation:} Current approaches break down KB results by flattening them into (\textit{subj-rel-obj}) triples. However, converting KB results into triples leads to loss of relationship amongst attributes in the result set. This makes the reasoning over memory difficult as model now has to infer relationships when retrieving values from memory. Instead, we use a multi-level memory which keeps the natural hierarchy of results intact (without breaking them into triples). We  store the queries and their corresponding results and individual values at different {\em levels}. We first attend on the queries and then on the results for each query to identify which result the user is referring to. This also enables us to handle user requests that refer to results from a previously executed query. We propose that a representation of all the values in the result, and not just one of the values (designated as \textit{subj}), should be used while attending over a result in KB. We attend on this {compound} representation of the result before attending on the individual key-value pairs in each result, to determine which value to copy into the generated  response.

\subsubsection{{Memory Representation}}
 
Let $q_1,q_2,...q_k$ be the queries fired to the knowledge base  till the current state of the dialogue. Every query $q_i$ is a set of key-value pairs $\{k^{q_i}_a : v_a^{q_i}, 1<a<n_{q_i}\}$, corresponding to the query's slot and argument where $n_{q_i}$ is the number of slots in query $q_i$.  For example, after the user utterance at Turn 3 in Table \ref{tab:example}, 
the query fired by the system on the knowledge base would be {\it \{'origin':'Dallas','destination':'Manheim','Start': 'Aug 26', 'end': 'Aug 31', 'Adults':1\}}. 
The execution of a query on an external knowledge base, returns a set of results. 
Let $r_{ij}$ be the $j^{th}$ result of query $q_i$. Each result $r_{ij}$ is also a set of slot-value pairs $\{k^{r_{ij}}_a : v_a^{r_{ij}}, 1<a<n_{r_{ij}}\}$ where $n_{r_{ij}}$ is the number of attributes in result $r_{ij}$. A visualization of the memory with queries and their corresponding results can be seen in Figure \ref{fig:memory-rep}.

The first level of memory contains the query representations. Each query $q_i$ is represented by $q^v_i$ = Bag of words over the word embeddings of values ($v_a^{q_i}$) in $q_i$. The second level of memory contains the result representations. Representation of each result $r_{ij}$ is given by $r^v_{ij}$ = Bag of words over the word embeddings of values ($v_a^{r_{ij}}$) in $r_{ij}$. The third level of memory contains the result cells which have the key-value pairs ($k^{r_{ij}}_a : v_a^{r_{ij}}$) of the results. 
The values ($v_a^{r_{ij}}$) which are to be copied into the system response are thus present in the final level of memory . 
We now describe how we apply attention over the context and multi-level memory. 

\subsection{Decoder} \label{sec:decoder}

The model generates the agent response word-by-word; a word at time step
$t$ is either generated from the decode vocabulary or is a value copied from one of the two memories (knowledge base or context memory). A soft gate $g_1$ controls whether a value is generated from vocabulary or copied from memory. Another gate $g_2$ determines which of the two memories is used to copy values.  
\subsubsection{Generating words:}
Let the hidden state  of the decoder at time $t$ be $h_t$.
\begin{equation}
h_t = GRU(\phi^{emb}(y_{t-1}),h_{t-1}) 
\end{equation}

The hidden state $h_t$ is used to apply attention over the input context memory. Attention is applied over the hidden states of the input bi-directional (BiDi) GRU encoder using the ``{\it concat}'' scheme as given in \cite{luong2015effective}. The attention for the $j$th word in the $i$th utterance is given by:
\begin{equation}
a_{ij}=\frac{exp(w_1^Ttanh(W_2tanh(W_3[h_t,h^e_{ij}])))}{\sum_{ij}exp(w_1^Ttanh(W_2tanh(W_3[h_t,h^e_{ij}])))}
\end{equation}
\textcolor{black}{The attention scores $a_{ij}$ are combined to create an attended context representation $d_t$}, 
\begin{equation}
d_t = \sum_{i,j}a_{i,j}h^e_{ij}
\end{equation}
and similar to \cite{luong2015effective}, the decoder word-generation distribution is given by  :
\begin{equation}
P_g(y_t) = softmax(W_1[h_t,d_t] + b_1)
\end{equation}
\subsubsection{Copying words from context memory:}
The input context memory is represented using the hidden states $h^e_{ij}$ of the input Bi-Di GRU encoder. Similar to \cite{gulcehre2016pointing}, the attention scores $a_{ij}$, are used as the probability scores to form the copy distribution $P_{con}(y_t)$ over the input context memory. 
\begin{equation}
P_{con}(y_t = w) = \sum_{ij:w_{ij}=w}a_{ij}
\end{equation}

\subsubsection{Copying entries from KB memory:}

The context representation $d_t$, along with the hidden state of decoder $h_t$, is used to attend over the multi-level memory. The first level attention, $\alpha_.$, is applied over the queries $q_.$. 
\begin{equation}
\alpha_i = \frac{exp(w_2^Ttanh(W_4[d_t,h_t,q_i^v]))}{\sum_iexp(w_2^Ttanh(W_4[d_t,h_t,q_i^v]))}
\end{equation}
The second level attention, $\beta_{i.}$, is the attention over the results $r_{i.
}$ of query $q_i$.

\begin{equation}
\beta_{ij} = \frac{exp(w_3^Ttanh(W_5[d_t,h_t,r^v_{ij}]))}{\sum_jexp(w_3^Ttanh(W_5[d_t,h_t,r^v_{ij}]))}
\end{equation}

 The product of first level attention and second level attention is the attention over results of all the queries in the multi-level memory. The weighted sum of the first level attention, second level attention and result representations gives us the attended memory representation, $m_t$. 

\begin{equation}
m_t = \sum_i\sum_j \alpha_i \beta_{ij} r^v_{ij}
\end{equation}

Each result is further composed of multiple result cells. On the last level of memory, which contains the result cells, we apply key-value attention similar to \cite{KVRet}. The key of the result cell is the word embedding of the slot, $k_a^{r_{ij}}$, in the result. The attention scores, $\gamma_{ij.}$, for the keys represent the attention over the result cells of each result $r_{ij}$. 

\begin{equation}
\gamma_{ijl} =  \frac{exp(w_4^Ttanh(W_6[d_t,h_t,\phi^{emb}(k_l^{r_{ij}})]))}{\sum_l exp(w_4^Ttanh(W_6[d_t,h_t,\phi^{emb}(k_l^{r_{ij}})]))}
\end{equation}


The product of first level attention $\alpha_i$, second level attention $\beta_{ij}$ and third level attention $\gamma_{ijl}$ gives the final attention score of the value $v_l^{r_{ij}}$ in the KB memory. These final attention scores when combined (Eq. \ref{eq:kb_dist}), form the copy distribution, $P_{kb}(y_t)$, over the values in KB memory.


\begin{equation} \label{eq:kb_dist}
P_{kb}(y_t=w) = \sum_{ijl:v_l^{r_{ij}}=w}\alpha_i \beta_{ij} \gamma_{ijl}
\end{equation}

\subsubsection{Decoding}
Similar to \cite{gulcehre2016pointing}, we combine the generate and copy distributions - we use gate $g_2$  (Eq. \ref{eq:g1}) to obtain the copy distribution $P_c(y_t)$  (Eq. \ref{eq:P-copy}) by combining $P_{kb}(y_t)$ and $P_{con}(y_t)$. 
\begin{equation} \label{eq:g1}
g_2 = sigmoid(W_7[h_t,d_t,m_t] + b_2)
\end{equation}
\begin{equation} \label{eq:P-copy}
P_c(y_t) = g_2P_{kb}(y_t) + (1-g_2)P_{con}(y_t)
\end{equation}

Finally, we use gate $g_1$ to obtain the final output distribution $P(y_t)$, by combining generate distribution $P_g(y_t)$ and copy distribution $P_c(y_t)$ as shown below:
\begin{equation}
g_1 = sigmoid(W_8[h_t,d_t,m_t] + b_3)
\end{equation}
\begin{equation}
P(y_t) = g_1P_{g}(y_t) + (1-g_1)P_{c}(y_t)
\end{equation}
We train our model by minimizing the cross entropy loss $-\sum_{t=1}^T log(P(y_t))$.

%% file: experiments.tex
\section{Experiments}
\subsection{Datasets}
We present our experiments using three real world publicly available multi-turn task oriented dialogue datasets: the InCar assistant \cite{KVRet}, CamRest \cite{CamRest} and the Maluuba Frames dataset \cite{elasri2017frames}. All three datasets contain human-human task oriented dialogues which were collected in a Wizard-of-Oz \cite{wen2016network} setting.

\noindent (i) \textbf{InCar assistant dataset} consists of $3031$ multi-turn dialogues in three distinct domains: calendar scheduling, weather information retrieval, and point-of-interest navigation. Each dialogue has it's own KB information provided and thus, the system does not have to make any queries. 

\noindent (ii) \textbf{CamRest dataset}, consists of $676$ human-to-human dialogues set in the restaurant reservation domain. There are three queryable slots (food, price range, area) that users can specify. 
\textcolor{black}{This dataset has currently been used for evaluating slot-tracking systems. \textcolor{black} Recent work by {\cite{lei2018sequicity} uses an end-to-end network without a KB and substitutes slot values with placeholders bearing the slot names in agent responses.} However, we formatted the data to evaluate end-to-end systems by adding API call generation from the slot values so that restaurant suggestion task can proceed from the KB results.} 

\noindent (iii) \textbf{Maluuba Frames dataset}, consists of $1369$ dialogues developed to study the role of memory in task oriented dialogue systems. The dataset is set in the domain of booking travel packages which involves flights and hotels. 
In contrast to the previous two datasets, this dataset contains dialogs that require the agent to remember all information presented previously as well as support results from multiple queries to the knowledge base. 
A user's preferences may change as the dialogue proceeds, and can also refer to previously presented queries (non-sequential dialog). \textcolor{black}{Thus, to store multiple queries, we require $3$ levels in our multi-level memory as compared to $2$ levels in the other datasets, since they don't have more than one query.} 
We do not use the dialogue frame annotations and use only the raw text of the dialogues. We map ground-truth queries to API calls that are also required to be generated by the model. \textcolor{black}{Recent work has used this dataset only for frame tracking \cite{schulz2017frame} and dialogue act prediction \cite{peng2017composite,tang2018subgoal}. To the best of our knowledge we are the first to attempt the end-to-end dialog task using this dataset.} Table \ref{tab:statistics} summarizes the statistics of the datasets.



\begin{table}[ht]
\scriptsize
\centering
\begin{tabular}{|c|c|c|c|}
\hline
&InCar&CamRest& Maluuba Frames\\
\hline
Train Dialogs&2425 &406 & 1095\\
\hline
Val Dialogs &302 &135 & 137 \\
\hline
Test Dialogs &304 &135 & 137 \\
\hline
Avg. no. of turns &2.6 &5.1 & 9.4\\
\hline
Avg length. of sys. resp. &8.6 &11.7 & 14.8 \\
\hline
Avg no. of sys. entities &1.6 & 1.7& 2.9\\
\hline
Avg no. of queries &0 & 1& 2.4\\
\hline
Avg no. of KB entries &66.1 & 13.5& 141.2 \\
\hline
\end{tabular}
\caption{Statistics for $3$ different datasets.} 
\label{tab:statistics}
\end{table}


 

\subsection{KB API Call Generation}

In this section, we briefly describe how the knowledge base queries are generated as API calls as part of the model response. The InCar assistant dataset has a fixed KB for each dialogue whereas the CamRest and Maluuba datasets require queries to be fired on a global KB. Queries in CamRest dataset can have 3 slots namely cuisine, area and pricerange, whereas those in Maluuba can have 8 slots, which are destination, origin, start date, end date, budget, duration, number of adults and children. Any query that is to be fired on the KB is expected to be generated by the model as an API call, by considering a fixed ordering of slots in the generated response. For eg., in CamRest dataset, ApiCall(area=south, pricerange=cheap) would be generated by the model as \textit{api\_call dontcare south cheap}, with \textit{dontcare} meaning that the user does not have any preference for cuisine and, \textit{south}, \textit{cheap} being the user constraints for area and pricerange respectively. Therefore, the task of API call generation typically involves copying relevant entities that are present in dialog context.

\subsection{Training}

Our model is trained end-to-end using Adam optimizer \cite{kingma2014adam} with a learning rate of $2.5e^{-4}$. The batch-size is sampled from [8,16]. We use pre-trained Glove vectors \cite{pennington2014glove} with an embedding size of 200. The GRU hidden sizes are sampled from [128, 256]. We tuned the hyper-parameters with grid search over the validation set and selected the model which gives best entity F1.

\subsection{Evaluation Metrics}
\begin{table*}[ht]
\scriptsize
\centering
\begin{tabular}{l||ccccc||ll||ll}
\hline
                                                                & \multicolumn{5}{c}{InCar}                                                                                                                 & \multicolumn{2}{c}{CamRest} & \multicolumn{2}{c}{Maluuba Frames }  \\ 
\hline \hline
Model                                                           & BLEU & F1 & \begin{tabular}[c]{@{}l@{}}Calendar\\~ ~~ F1\end{tabular} & \begin{tabular}[c]{@{}c@{}}Weather \\F1\end{tabular} & \begin{tabular}[c]{@{}c@{}}Navigate \\F1\end{tabular} & BLEU & F1                    & BLEU & F1                           \\ 
\hline
Attn seq2seq  \cite{luong2015effective}                                                  & 11.3      & 28.2  &  36.9                                                &     35.7                                                 &   10.1          & 7.7     &  25.3                     &    3.7  &  16.2                          \\
Ptr-UNK    \cite{gulcehre2016pointing}                                                     &      5.4& 20.4   &     22.1                                                      &                     24.6                                 &             14.6&  5.1    &    40.3                    &5.6     &                  25.8            \\
KVRet   \cite{KVRet}                                                        & 13.2    & 48.0   &                62.9                                           &        47.0                                              & 41.3             & 13.0    &   36.5                    &  10.7    & 31.7                             \\
Mem2Seq            \cite{Mem2Seq}                                             &  11.8    & 40.9   &      61.6                                                     & 39.6                                                     & 21.7            &   14.0   &    52.4                   & 7.5     &      28.5                        \\ 
\hline
Multi-level Memory Model (MM)  & {\bf 17.1}   & {\bf 55.1}     & {\bf 68.3}   & {\bf 53.3}                                                         &    {\bf 44.5}                                                 & {\bf 15.9}            & {\bf 61.4}     & {\bf 12.4}                      &   {\bf 39.7}                                \\    \hline                  
\end{tabular}
\caption{Comparison of our model with baselines} \label{tab:results}
\vspace{-0.3cm}
\end{table*}
\subsubsection{BLEU}

We use the commonly used BLEU metric \cite{papineni2002bleu} to study the performance of our systems as it has been found to have strong correlation \cite{sharma2017relevance} with human judgments in task-oriented dialogs. 
\subsubsection{Entity F1}
To explicitly study the behaviour of different memory architectures, we use the entity $F1$ to measure how effectively values from the knowledge base are used in the dialog. To compute the entity F1, we micro-average the precision and recall over the entire set of system responses to compute the micro F1\footnote{We observe that \cite{Mem2Seq} reports the micro average of recall as the micro F1.}. 
For the InCar Assistant dataset, we compute a per-domain entity F1 as well as the aggregated entity F1. Since our model does not have slot-tracking by design, we evaluate on entity F1 instead of the slot-tracking accuracy as in \cite{henderson2014second,wen2016network}
\subsection{Baselines}
We experiment with the following baseline models for comparing the performance of our Multi-Level Memory architecture:

\noindent{\bf Attn seq2seq}\footnote{We use the implementation provided by \cite{Mem2Seq} at https://github.com/HLTCHKUST/Mem2Seq\label{baselines}} \cite{luong2015effective}: A model with simple attention over the input context at each time step during decoding.

\noindent\textbf{Ptr-UNK}\textsuperscript{\ref{baselines}} \cite{gulcehre2016pointing}: The model augments a sequence-to-sequence architecture with attention-based copy mechanism over the encoder context. 

\noindent\textbf{KVRet} \cite{KVRet}: The model uses key value knowledge base in which the KB is represented as triples in the form of $subject-relation-object$. This model does not support copying words from context. The sum of word embeddings of $subject$, $relation$ is used as the key of the corresponding $object$. 

\noindent\textbf{Mem2Seq}\textsuperscript{\ref{baselines}} \cite{Mem2Seq}: The model uses a memory networks based approach for attending over dialog history and KB triples. 
During decoding, at each time step, the hidden state of the decoder is used to perform multiple hops over a single memory which contains both dialog history and the KB triples to get the pointer distribution used for generating the response. 

\subsection{Results}

Table \ref{tab:results} shows the performance of our model against our baselines. We find that our multi- level memory architecture comprehensively beats all existing models, thereby establishing new state-of-the- art benchmarks on all three datasets. Our model outperforms each baseline on both BLEU and entity F1 metrics.

\noindent {\bf InCar: } On this dataset, we show entity F1 scores for each of the scheduling, weather and navigation domains. Our model has the highest F1 scores across all the domains. It can be seen that our model strongly outperforms Mem2Seq on each domain. A detailed study reveals that the use of triples cannot handle cases when a user queries with {\it non-subject} entries or in cases when the response requires inferencing over multiple entries. In contrast, our model is able to handle such cases since we use a compound representation of entire result (bag of words over values) while attending on that result. 

\noindent {\bf CamRest: } 
Our model achieves the highest BLEU and entity F1 scores on this dataset. From Table \ref{tab:results}, we see that simpler baselines like Ptr-UNK show competitive performance on this dataset because, as shown in Table \ref{tab:statistics}, CamRest dataset has relatively fewer KB entries. Thus, a simple mechanism for copying from context results in \textcolor{black}{good} entity F1 scores.

\begin{table}[ht]
\centering
\scriptsize
\begin{tabular}{c||ll|ll|ll}
\hline
                                &     \multicolumn{2}{c|}{\bf InCar}       &     \multicolumn{2}{c|}{\bf CamRest} & \multicolumn{2}{c}{\bf Maluuba}    \\
\cline{2-7}
                                & Ctxt.       & KB & Ctxt. & KB   & Ctxt. & KB  \\ \hline \hline

Mem2Seq                         &    66.2          &   25.3      &       63.7         &  36.5       &   17.7      &       8.9          \\
Multi-level Mem. &      81.6       &       37.5  &         70.1       & 53.4        &    27.2     &  14.6           
\\
\hline
\end{tabular}
\caption{Percentage (\%) of category-wise (context vs KB) ground truth entities correctly captured in generated response. Abbreviation Ctxt denotes context.}
\label{tab:entitywise}
\end{table}

\begin{table*}[ht]
\scriptsize
\centering
\begin{tabular}{l||ccccc||ll||ll}
\hline
                                                                & \multicolumn{5}{c}{InCar}                                                                                                                 & \multicolumn{2}{c}{CamRest} & \multicolumn{2}{c}{Maluuba Frames}  \\ 
\hline \hline
Model                                                           & BLEU & F1 & \begin{tabular}[c]{@{}l@{}}Calendar\\~ ~~ F1\end{tabular} & \begin{tabular}[c]{@{}c@{}}Weather \\F1\end{tabular} & \begin{tabular}[c]{@{}c@{}}Navigate \\F1\end{tabular} & BLEU & F1  & BLEU & F1                                            \\ 
\hline
Unified Context and KB memory (Mem2Seq) &   11.8   &  40.9  & 61.6                                                         &    39.6                                                &21.7      &  14.0  &  52.4 & 7.5 & 28.5                                             \\ \hline

Separate Context and KB Memory &   14.3   &  44.2  & 56.9                                                         &    54.1                                                &24.0      &  14.3  &  55.0 & 12.1 & 36.5                                           \\
+Replace KB Triples with Multi-level memory & 17.1    & {55.1}     & 68.3   & {53.3}                                                         &    {44.5}                                                 & 15.9           & 61.4       & 12.4 & 39.7                              \\                 
\hline
\end{tabular}
\caption{Model ablation study : Effect of (i) separate memory and (ii) multi-level memory design.} \label{tab:ablation}
\vspace{-0.3cm}
\end{table*}

\noindent{\bf Maluuba Frames: }The Maluuba Frames dataset was \textcolor{black}{introduced} for the frame tracking task. Here, 
a dialog frame is a structured representation of the current dialog state. 
Instead of explicitly modeling the dialog frames, we use the context representation $d_t$ to directly attend on the Multi-level memory. As Table \ref{tab:statistics} shows, this dataset contains significantly longer contexts as well as larger number of entities, as compared to the other two datasets. In addition, unlike other datasets, it also contains non-linear dialog flows where a user may refer to previously executed queries and results. The complexity of this dataset is reflected in the relatively lower BLEU and F1 scores as compared to other datasets.
\subsection{Analysis}
\subsubsection{Entity source-wise performance} 
To further understand the effect of separating context memory from KB memory and using a multi-level memory for KB,  Table \ref{tab:entitywise} shows the percentage of ground-truth entities, according to their category, which were also present in the generated response. For example, on the InCar dataset, out of the $930$ entities in ground-truth response that were to be copied from the KB, our model was able to copy $37.5\%$
 of them into the generated response. From Table \ref{tab:entitywise}, it can be seen that our model is able to correctly copy a significantly larger number of entities from both, KB and context, as compared to the recent Mem2Seq model in all datasets. 

\subsubsection{Model ablation study}
We report results from ablation studies on all three datasets. Table \ref{tab:ablation} shows the incremental benefit obtained from individual components used in our model. We investigate the gains made by (i) Using separate memory for context and KB triples (ii) Replacing KB triples with a Multi-level memory. We use the recent Mem2Seq model for comparison with a unified context and KB memory model. 

As can be seen from Table \ref{tab:ablation}, the separation of context memory and KB memory leads to a significant improvement in BLEU and F1 scores on all datasets. This validates our hypothesis that storing context words and KB results in a single memory confuses the memory reader. 
The use of a multi-level memory instead of triples leads to further gains. This suggests, better organization of KB result memory by keeping the natural hierarchy intact is beneficial.

\subsubsection{Error Analysis}
We analyzed the errors made by our dialog model on 100 dialog samples in test set of Maluuba Frames. We observed that the errors can be divided into five major classes: (i) Model outputs wrong KB result entry due to incorrect attention (27\%), (ii) Model returns package details instead of asking for more information from the user (16\%), (iii) Model incorrectly captures user intent (13\%), (iv) Model makes an error due to non-sequential nature of dialog (22\%). 
In such errors, our model either generates an API call for a result already present in memory, or our model asks for a query-slot value that was already provided by the user, (v) Data specific characteristics such as insufficient samples for certain classes of utterances (eg: more than one package returned) or returning different, but meaningful package attributes as compared to ground-truth data, contribute to 22\% of the errors.

\begin{table}[ht]
\centering
\scriptsize
\begin{tabular}{c||lll|lll}
\hline
&     \multicolumn{3}{c|}{\bf CamRest}       &     \multicolumn{3}{c}{\bf Maluuba}\\
\hline
& Info. & Lang. & MRR & Info. & Lang. & MRR \\
\hline
\hline
KVRet &  2.49 & 4.38 & 0.57 & 2.42 & 3.31&0.64 \\
\hline
Mem2Seq & 2.48 & 3.72 & 0.51 & 1.78& 2.55&0.50 \\
\hline
Our Model & {\bf 3.62}& {\bf 4.48}& {\bf 0.76}& {\bf 2.45}& {\bf 3.93}& {\bf 0.69} \\
\hline
\end{tabular}
\caption{Informational accuracy and language quality scores using a human study.}
\label{tab:human}
\end{table}

\subsection{Human Evaluation}
We also conducted a blind user study that compared outputs from our model, Mem2Seq and KVRet systems. We used $96$ randomly selected examples from each test split of Maluuba and CamRest datasets resulting in a total of 192 examples. Our study was split across 8 users who were provided with results fetched from the KB, current dialog context, gold response and the outputs of each of the models. Model outputs were shuffled in each example and users were asked to score each output between 1 (lowest) to 5 (highest) in terms of its accuracy of information in response and the quality of language. 
The results of this study are presented in Table \ref{tab:human}. We also report the MRR (mean-reciprocal rank) for model preference along with other scores. 

It can be seen that our model consistently ranks high for both information accuracy and language quality as well as reports a higher MRR. To further understand the quality of model performance, we asked the human evaluators whether their best ranked model output was a useful response. We saw that the evaluators agreed in 76.04\% and 58.33\% of the cases for CamRest and Maluuba datasets respectively. We observe that the results from human evaluation go hand-in-hand with automatic evaluation and reinforce our claim that separating context, KB memory and using a multi-level representation for the KB memory are useful for improving dialog modeling.

\begin{table}[ht]
\centering
\tiny
\begin{tabular}{p{8mm}p{2mm}p{54mm}}
\textbf{Role} & \textbf{Turn } & \textbf{Utterance} \\                                                                                                                                                                                                                
\hline \hline 
Agent & 1 & {hello ! how can i help you today ?}\\
\hline
User & 2 & \begin{tabular}[c]{@{}l@{}} {\it i just need to dream a bit. i have 4500 dollars but no vacation days.}\\ {\it i'm wondering what a hypothetical trip to Manas would be like}\end{tabular} \\
\hline
Agent & 2 & i can help with that, where are you departing from \\
\hline
User & 3 & \begin{tabular}[c]{@{}l@{}} {\it i'm in Tijuana, not departing any time soon unfortunately though}\end{tabular} \\
\hline
 ... & ... & ... \\
 Agent & 5 & \begin{tabular}[c]{@{}l@{}}\textcolor{blue}{\texttt{[API\_CALL manas tijuana 29/8 16/9 4500 ...]}}\end{tabular} \\
 ... & ... & ... \\
\hline
User & 7 & {\it that's fine, what about pittsburgh}\\
\hline
Agent & 7 & \begin{tabular}[c]{@{}l@{}} \textcolor{blue}{\texttt{[API\_CALL pittsburgh tijuana ... 4500 ...]}} \end{tabular}\\
\hline
 ... & ... & ... \\
Agent & 8 & \begin{tabular}[c]{@{}l@{}}that i do have! would you be satisfied with a 3.5 star hotel ?\end{tabular}\\
\hline
User & 9 & \begin{tabular}[c]{@{}l@{}}{\it as of now i'm satisfied with a tent and a backpack .}\\ {\it what are the hotel details?}\end{tabular} \\
\hline
Agent & 9 & \begin{tabular}[c]{@{}l@{}}the hotel is situated near a park, comes with free parking, wifi \\and breakfast ! does this sound good\end{tabular} \\
\hline
User & 10 & {\it sounds quite nice, what about the guest rating ?}\\
\hline
\hline
\textbf{Gold} & & \textcolor{black}{the guest rating is impressive : \textbf{8.86}}\\
\hline
\hline
\textbf{Mem2Seq} & & \textcolor{red}{this is the only available with the same} \\
\hline
\textbf{KVRet} & & \begin{tabular}[c]{@{}l@{}} \textcolor{red}{it's a 2.5 star hotel near a park, museum and airport.it has a}\\\textcolor{red}{guest rating of 8.22} \end{tabular} \\
\hline
\hline
\begin{tabular}[c]{@{}l@{}}\textbf{Our Model}\end{tabular}& & \begin{tabular}[c]{@{}l@{}}\textcolor{cadmiumgreen}{the onyx isle hotel has a \textbf{8.86} guest rating and offers free parking,}\\ \textcolor{cadmiumgreen}{breakfast and wifi.}\end{tabular}\\
\hline
\end{tabular}
\caption{Comparing the responses generated by various models on an example in test set of Maluuba Frames.} 
\label{tab:viz}
\end{table}

\subsection{Attention Visualization}

Analyzing the attention weights is a useful way to understand how the model is inferencing over the memory to copy entities from it. Table \ref{tab:viz} shows an example of a dialog from the Maluuba Frames dataset and the outputs generated by different models. Here, the user first wants to know about packages to `Manas' and then requests for trips to `Pittsburgh'. Later, the user becomes interested in the 3.5 star hotel in Pittsburgh which was suggested by the agent and wants to know its guest rating. It can be seen from Table \ref{tab:viz} that our model outputs the correct guest rating (8.86) of the hotel. Mem2Seq fails to understand the context and generates an irrelevant response. KVRet generates a readable response but points to the guest rating of a different hotel. 

\begin{figure}[ht]
	\centering
	\begin{subfigure}[b]{.48\textwidth}
		\centering
		\includegraphics[scale=0.36]{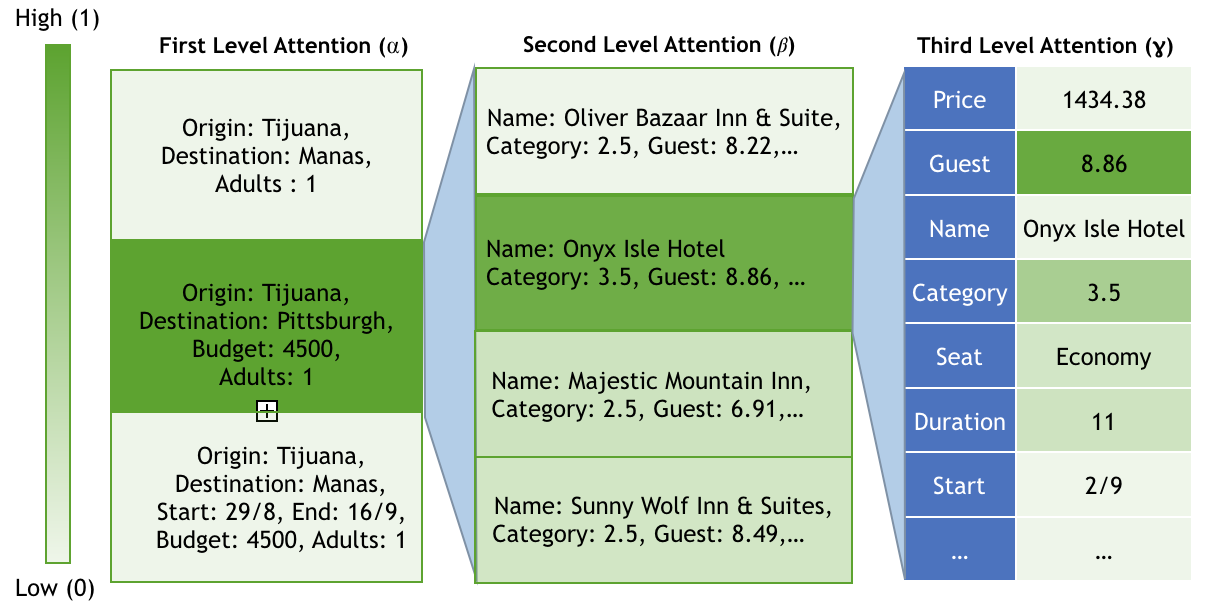}
	    \caption{{\fontsize{10}{12}\selectfont Attention over the multi-level KB memory}}\label{fig:kb-attn}
	\end{subfigure}
	\begin{subfigure}[b]{.50\textwidth}
		\centering
		\includegraphics[scale=0.4]{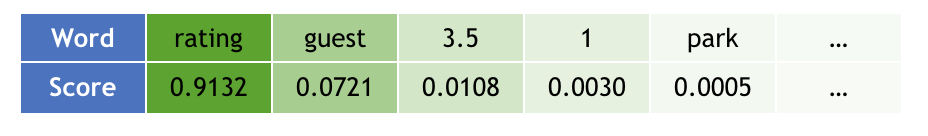}
	    \caption{{\fontsize{10}{12}\selectfont Decreasing order of attention scores over words in\\ dialogue context}}\label{fig:context-attn}
	\end{subfigure}
    \caption{Visualization of attention over memory while generating the word `8.86' for the example in Table \ref{tab:viz}.} 
    \label{fig:mem-viz}
\end{figure}

The attention over the memory while generating the word `$8.86$' for this example is shown in Fig \ref{fig:mem-viz}. 
Fig \ref{fig:kb-attn} shows that the query with destination as `Pittsburgh' gets the highest attention and among the results of this query, the package with the 3.5 star rated hotel gets highest attention. Within this result, the model gives highest score to the result cell with guest rating as the key. To further understand why the correct result hotel gets higher attention, Fig \ref{fig:context-attn} shows the attention scores over the words in context memory. The context representation $d_t$ captures the important words (3.5, guest, rating) in context which are in-turn used to apply attention over the multi-level memory.

Lastly, studying the values of the gates $g_1$ (prob. of generating from vocabulary) and $g_2$ (prob. of copying from KB), we found that gate $g_1$ had a probability value of \textit{0.08} thereby driving the model to copy from memory instead of generating from output vocabulary and gate $g_2$, with a probability value of \textit{0.99}, was responsible for selecting KB memory over context memory.

%% file: conclusion.tex
\section{Conclusion}
In this paper, we presented an end-to-end trainable novel architecture with multi-level memory for task oriented dialogues. Our model separates the context and KB memory and combines the attention on them using a gating mechanism. The multi-level KB memory reflects the natural hierarchy present in KB results. This also allows our model \textcolor{black}{to support non-sequential dialogs where a user may refer to a previously suggested result.} We find that our model beats existing approaches by 15-25\% on both entity F1 and BLEU scores, establishing state-of-the-art results on three publicly available real-world task oriented dialogue datasets. In a user study comparing outputs from our system against recent models, we found that our model consistently scored higher for both language quality as well as correctness of information in the response. 
We also present the benefits of each of our design choices by performing an ablation study. In future work, we would like to incorporate better modeling of latent dialog frames so as to improve the attention signal on our multi-level memory. As our error analysis suggests, nearly 22\% of the errors could possibly be reduced by improved modeling of the dialog context. We believe that model performance can also be improved by capturing user intent better in case of non-sequential dialog flow. 